\documentclass{aims}

\usepackage{txfonts}
\usepackage{booktabs}
\usepackage{array}
\usepackage{multirow}
\usepackage{color, soul}
\usepackage{cite}

\def\typeofarticle{Article}
\def\currentvolume{5}
\def\currentissue{x}
\def\currentyear{2018}

\def\ppages{xxx--xxx}
\def\DOI{}
\def\Received{}
\def\Accepted{}
\def\Published{}

\numberwithin{equation}{section}

\begin{document}

\title{Resampling detection of recompressed images via dual-stream convolutional neural network}

\author{%
  Gang Cao\affil{1,2,},
  Antao Zhou\affil{1},
  Xianglin Huang\affil{1,2}\corrauth,
  Gege Song\affil{1},
  Lifang Yang\affil{1,2} and
  Yonggui Zhu\affil{3}
}

\shortauthors{the Author(s)}

\address{%
  \addr{\affilnum{1}}{School of Computer Science and Cybersecurity, Communication University of China, Beijing 100024, China}
  \addr{\affilnum{2}}{Key Laboratory of Convergent Media and Intelligent Technology (Communication University of China), Ministry of Education, China}
  \addr{\affilnum{3}}{School of Data Science and Media Intelligence, Communication University of China, Beijing 100024, China}}
\corraddr{huangxl@cuc.edu.cn
}

\begin{abstract}
Resampling detection plays an important role in identifying image tampering, such as image splicing. Currently, the resampling detection is still difficult in recompressed images, which are yielded by applying resampling followed by post-JPEG compression to primary JPEG images. Except for the scenario of low quality primary compression, it remains rather challenging due to the widespread use of middle/high quality compression in imaging devices. In this paper, we propose a new convolution neural network (CNN) method to learn the resampling trace features directly from the recompressed images. To this end, a noise extraction layer based on low-order high pass filters is deployed to yield the image residual domain, which is more beneficial to extract manipulation trace features. A dual-stream CNN is presented to capture the resampling trails along different directions, where the horizontal and vertical streams are interleaved and concatenated. Lastly, the learned features are fed into Sigmoid/Softmax layer, which acts as a binary/multiple classifier for achieving the blind detection and parameter estimation of resampling, respectively. Extensive experimental results demonstrate that our proposed method could detect resampling effectively in recompressed images and outperform the state-of-the-art detectors.
\end{abstract}


\maketitle

\section{Introduction}

Digital images play an important part in delivering information, leading public opinion and taking evidence in legal proceedings and criminal investigations. Photo editing and operation tools make progress in creating forged images without leaving any perceptible artifacts. Malicious image manipulation may lead to serious ethical and legal problems. Therefore, it is significant to detect the image manipulation blindly \cite{Stamm2013Information}.

Image manipulation generally destroys some inherent consistency within original images, or leaves new unique operation trail. Such clues could be exploited to detect manipulation blindly by forensic analysts. In the last decade, numerous methods have been developed to detect various types of image manipulations, such as JPEG compression \cite{Farid2009Exposing,Luo2010JPEG}, contrast enhancement \cite{Cao2014Contrast,scis14,caee18}, sharpening \cite{Ding18}, median filtering \cite{Kang2013Robust}, image splicing \cite{Ye2007Detecting,Dirik2009Image,Ferrara2012Image,Amerini2014Splicing,Huh2018Fighting,Zhou2018Learning,Bunk2017Detection} and computer graphic rendering \cite{cg2018}.

Blind detection of image resampling has received extensive attention. Existing image resampling detection methods operate by identifying operation traces from spatial \cite{Popescu2005Exposing,V2015An,Vazquez2017A,Ta2012Effective,Mahdian2008Blind,Qiao2018} and frequency domains \cite{Feng2011An,Kirchner2008Fast}. Popescu \emph{et al.} propose to detect the statistical correlation of interpolated pixels by expectation maximization (EM) algorithm \cite{Popescu2005Exposing}. Mahdian and Saic propose to detect pixel interpolation traces by capturing the statistical changes on signal covariance structure \cite{Mahdian2008Blind}. In \cite{Vazquez2017A}, subspace decomposition and random matrix theory are used to identify resampling traces in upscaled images. Feng \emph{et al.} exploit normalized energy density to derive a 19-dimensional feature vector, which is fed into a SVM classifier \cite{Feng2011An}. In addition, Kirchner \emph{et al.} \cite{Kirchner2008Fast} model the detectable resampling artifacts in spatial and frequency domains by means of the variance of prediction residue. A semi-intrusive forensic method is proposed to identify the category of resampling operators in \cite{fsi12}. Pasquini \emph{et al.} \cite{Pasquini2019Identification} analyze the detectability of downsampling in theory.

However, such prior resampling detection methods are rather fragile with post-JPEG compression, even for the quality factor Q around 95 \cite{Bunk2017Detection,Bayar2017On}. The periodic pattern introduced by resampling is disturbed by JPEG blocking artifacts.
In order to detect resampling in JPEG images, the periodicity in the second derivative of interpolated images is employed as evidence of resampling detection \cite{Gallagher2005Detection}. Furthermore, Nataraj \emph{et al.} \cite{Nataraj2009Adding} propose to suppress JPEG periodic patterns by adding Gaussian noise to post-compressed images. However, such prior methods are ineffective in low quality JPEG images. Kirchner and Gloe \cite{Kirchner2009On} demonstrate that the previous JPEG compression before resampling can well detect resampling in recompressed images.
Although such an approach can detect resampling in the cases of low and moderate quality primary compression, it is invalidated for the high quality case. In \cite{CHEN20178}, the operation chain including JPEG compression and resampling is detected by the transformed block artifacts extracted from pixel and discrete cosine transforms (DCT) domains. Li \emph{et al.} propose a data-driven resampling detection method based on improved spatial rich model (SRM) features \cite{Haodong2018Identification}.

Convolutional neural network (CNN) is an important deep learning algorithm widely used in computer vision. A constrained convolutional neural network has been proposed for detecting image resampling in the prior works \cite{Bayar2017On,Bayar2018Constrained}. Such a method achieves the best performance in resampling detection compared with previous methods.

In this paper, we propose a new CNN framework to detect resampling in recompressed images. A noise extraction layer consisting of high pass filters is inserted into the CNN architecture. An input image can be converted into noise residuals by such filters for suppressing the influence of image contents. Meanwhile, the dual-stream CNN is employed to extract the high level feature of resampling traces. Note that resampling is often enforced locally to adjust region dimensions for creating realistic spliced images. Additionally, the proposed dual-stream CNN method is applied to the detection of such resampling-involved image splicing. Extensive experimental results verify that the proposed resampling detection method can achieve the state-of-the-art performance.

\begin{figure}
\begin{center}
\includegraphics[scale=0.11]{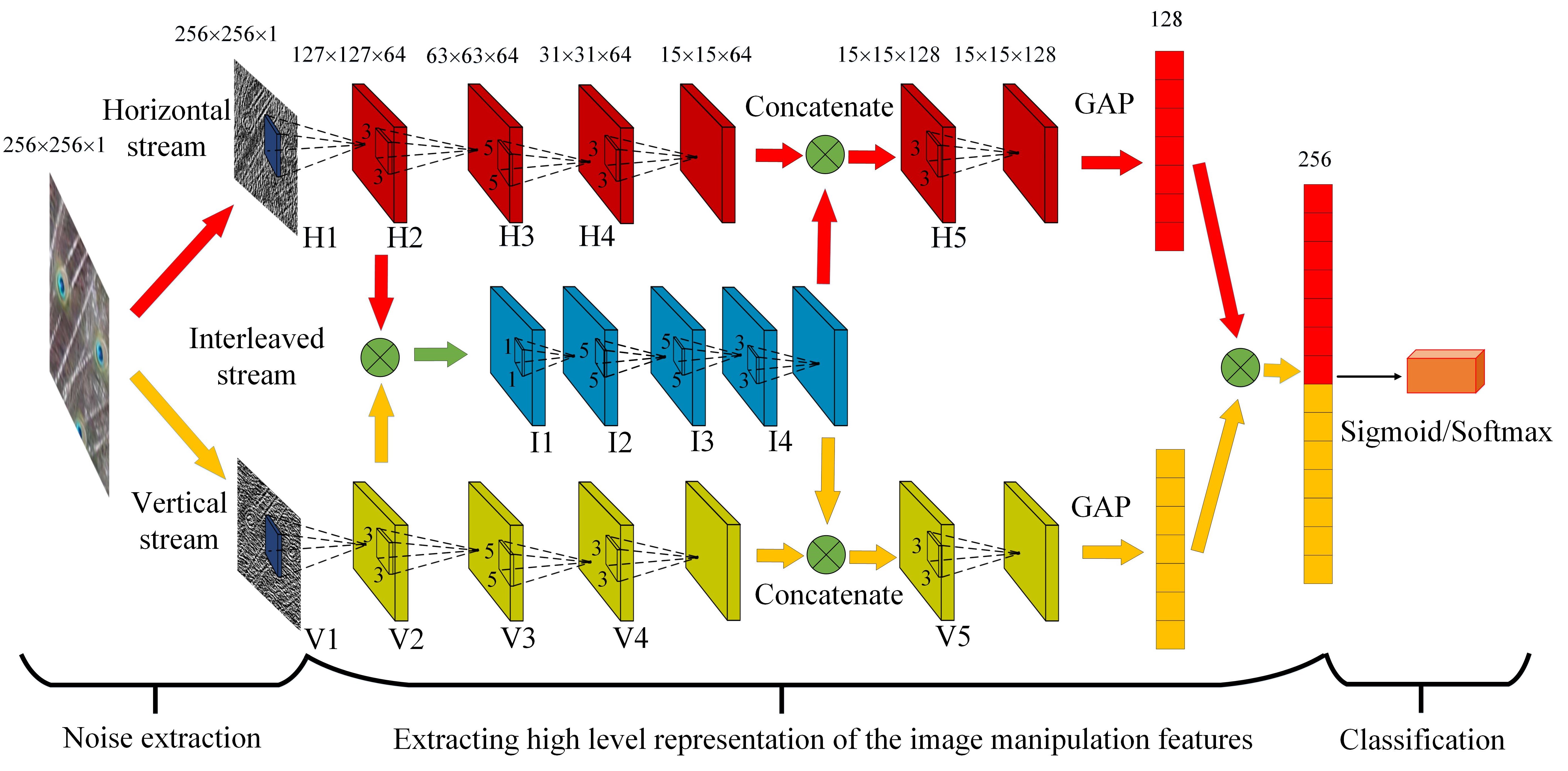}
\caption{Proposed dual-stream CNN-based resampling detector. H1-H5, I1-I4 and V1-V5 denote the convolution layers of horizontal, interleaved and vertical streams, respectively. GAP means global average pooling.}
\label{Fig1}
\end{center}
\end{figure}

In this work, we have made the following major contributions:
\begin{itemize}
\item A noise extraction layer is constructed by using two low-order high pass filters to suppress image content and highlight operational features.
\item A dual-stream convolutional neural network is proposed based on the methodology of two streams feature fusion, which could automatically learn the resampling features in images.
\end{itemize}

The rest of the paper is organized as follows. Section 2 describes the proposed resampling detection scheme in detail. Experimental results and discussions are presented in Section 3, followed by the conclusion drawn in Section 4.

\section{Proposed Resampling Detection Scheme}
In this section, the proposed image resampling detection scheme is presented in detail.
Firstly, we describe the overall framework of the network in subsection 2.1, and then introduce the design motivation and function of each module in subsections 2.2-2.4, respectively. Finally, the specific functions of each layer in the dual-stream CNN framework are analyzed in subsection 2.5.
\begin{figure}
\begin{center}
\includegraphics[scale=0.4]{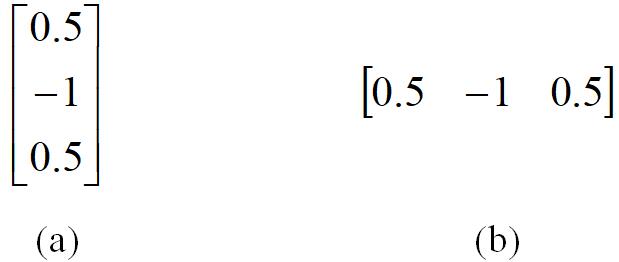}
\caption{Low-order highpass filters used in the noise extraction layer of (a) horizontal and (b) vertical streams.}
\label{Fig2}
\end{center}
\end{figure}

\subsection{Overview of proposed scheme}
We address resampling detection as a pattern classification problem, which would be resolved by deep learning methodology.
Figure 1 depicts the overall design of our proposed dual-stream CNN architecture for resampling detection.
It consists of three main components. $First$, in order to capture different dimensional features, the adjacent pixels differences from both horizontal and vertical directions are extracted by a noise extraction layer. $Second$, the high level representation of image manipulation features are generated by horizontal and vertical streams. In order to capture the correlation of both directions, horizontal and vertical streams are interleaved and concatenated. $Lastly$, the concatenated features of horizontal and vertical streams are fed into the sigmoid/softmax layer.
Note that the sigmoid layer is used for binary classification, and the softmax layer is used for multiple classification. Below we present a detailed introduction of such three components and the different layers used in our CNN architecture.

\subsection{Noise extraction}
Resampling detection is typically interfered by image content which should be suppressed.
It has been found that the resampling trace generally exists in the redundant domain of spatial-domain images and is irrelevant to image content \cite{Zhou2018Learning,Fridrich2012Rich,Ma2019,Zhang2018,Luo2016,Chen2017Image}.
In our approach, noise is modeled by the residual between a pixel and its estimation yielded by interpolating only neighboring pixels.
To accomplish this, a special convolution layer is set as the first layer of the network, namely noise extraction layer.
As shown in Figure 2, two low-order highpass filters are chosen as convolution kernels of noise extraction layers, which would not need to be trained cautiously.
The adjacent pixels differences in horizontal and vertical directions can be captured by the filters shown in Figures 2(a) and (b), respectively.
In particular, a grayscale image of $256\times256$ pixels is first convolved with $1\times3$ and $3\times1$ filters with stride 1 and padding 1.
Such filters could learn noise features from windows through the correlation of local pixels.
As a result, the noise extraction layers would yield a noise map of prediction residuals with the dimension of $256\times256\times1$.

\subsection{Feature extraction}
\subsubsection{Horizontal and vertical streams}
Noise from horizontal and vertical directions is extracted through the noise extraction layer.
Then high-level representation of resampling trace would be extracted from such noise.
Experimental observations show that a part of evidence would be lost and the overall resampling detection performance would degrade if a single, instead of double, direction(s) of noise gradient was used.
Therefore, two identical pipelines (horizontal stream and vertical stream) are designed to extract features independently from different directions, and the weights of two pipelines are not shared.
As shown in Figure 1, the horizontal and vertical streams are composed of five similar groups.
Each group consists of four layers including the convolution, batch normalization, activation and pooling ones.
The fifth group receives additional features from the interleaved stream. Finally, the features generated by the horizontal/vertical and interleaved streams are concatenated.

\subsubsection{Interleaved stream}
In order to determine the resampling behavior, it requires to fuse the two directional features for making a decision together. Therefore, we propose a novel feature fusion strategy of interleaved stream, which consists of four similar groups. The first group consists of a convolutional layer, a batch normalization layer and an activation layer. The remaining groups have an additional pooling layer. The features from the first group in horizontal and vertical streams are concatenated, and then enforced by $1\times1$ convolutional kernels with stride 1. Such a type of convolution kernel weights each position of the two feature graphs linearly for fusion. Then the remaining three groups further extract the high-level representation of fused features. Finally, the feature map output from interleaved stream is interpolated back to the horizontal and vertical streams. As cooperative learning, the horizontal and vertical streams are aware of each other without affecting the feature extraction in each direction.

\subsection{Classification}
The final features learned from the previous layers are fed into the classifier, which is a fully connected layer using softmax or sigmoid functions.
Through such a classifier, the probability that the feature belongs to each category is obtained, and the most probable category is the result of the classifier. The two involved functions are specified by

\begin{equation}
\tag{1}
    P(y=1|x) = \frac{1}{1 + e^{-z}}
\end{equation}

\begin{equation}
\tag{2}
    P(y=j|x) = \frac{e^{z_j}}{\sum_{k = 0} ^Ke^{z_k}}
\end{equation}
where Eq. (1) is a sigmoid function used in binary classification and $z$ is the value of neurons in the full connected layer. $P(y=1|x)$ is the probability that $x$ belongs to the positive category. Eq. (2) is a softmax function used in multiple classification, where $z_j$ is the value of the $j^{th}$ neuron in the full connected layer. $P(y=j|x)$ belongs to the probability that $x$ is the $j^{th}$ category.

\subsection{Layer architectures}
\subsubsection{Convolution layer}
The convolution layer in CNN is used to extract features. The input feature maps in convolution layers are computed as

\begin{equation}
\tag{3}
    F^{(n)}_j = \sum_{k = 0} ^K F^{(n-1)}_k * \omega^{(n)}_{kj} + b^{(n)}_j
\end{equation}
where $*$ indicates a 2d convolution operation, $F^{(n-1)}_k$ signifies the $k^{th}$ feature map generated in the $(n-1)^{th}$ layer, $\omega^{(n)}_{kj}$ indicates the $k$ channel of the $j^{th}$ convolution kernel in the $n^{th}$ layer, $b^{(n)}_j$ is the $j^{th}$ bias in the $n^{th}$ layer and $F^{(n)}_j$ is the $j^{th}$ feature map generated in the $n^{th}$ layer.

Three convolution kernels with different sizes ($1\times1$, $3\times3$ and $5\times5$) are used in the proposed model. In horizontal/vertical stream, the convolution layers ``H/V1$\sim$5'' are with 64, 64, 64, 64, 128 filters of size $5\times5\times1$, $3\times3\times64$, $5\times5\times64$, $3\times3\times64$ and $3\times3\times128$, respectively. The output dimensions of such convolutional layers are $256\times256\times64$, $127\times127\times64$, $63\times63\times64$, $31\times31\times64$, $15\times15\times64$, respectively. In interleaved stream, the convolution layers ``I1$\sim$4'' are with 64 filters of size $1\times1\times128$, $5\times5\times64$, $5\times5\times64$, $3\times3\times64$, respectively. Their output dimensions are $127\times127\times64$, $127\times127\times64$, $63\times63\times64$, $31\times31\times64$, respectively. The stride is set as 1 in all convolution layers.

\subsubsection{Batch normalization layer}
In order to solve the change of the data distribution in the middle layer during the training of CNN, we need to normalize the feature maps generated by the convolutional layers. To do this, a batch normalization layer is used between the convolution layer and the activation layer. The batch normalization operation within the CNN architecture is given from Eqs. (4) to (7).

First, the mean and variance of all data in a batch are calculated as
\begin{equation}
\tag{4}
    \mu = \frac{1}{m}\sum_{i = 0} ^mx_i
\end{equation}
\begin{equation}
\tag{5}
    \sigma^2 = \frac{1}{m}\sum_{i = 0} ^m(x_i - \mu)^2
\end{equation}
where $m$ denotes the number of data in a batch, $x_i$ signifies the $i^{th}$ data in a batch, $\mu$ and $\sigma^2$ indicate the mean and variance, respectively. Then each data is normalized to generate a new data $\hat{x}_i$ with a mean of 0 and a variance of 1. That is,
\begin{equation}
\tag{6}
    \hat{x}_i = \frac{x_i - \mu}{\sqrt{\sigma^2 + \epsilon}}\\
\end{equation}
where $\epsilon$ is small floating-point number greater than 0 to prevent dividing by zero errors. Finally, all data is scaled and shifted as
\begin{equation}
\tag{7}
    y_i = \gamma\hat{x}_i + \beta
\end{equation}
where $y_i$ indicates the $i^{th}$ output of the batch normalization layer. $\gamma$ and $\beta$ are the parameters learned by a network.

The newly generated data can make better use of the non-linear function of the activation function. Huge change in back layer parameters caused by small change in front layer parameters is prevented by batch normalization layer.

\subsubsection{Activation layer}
In order to compensate for the expressive deficiency of linear models, the convolution layer is generally followed by an activation layer containing a non-linear function called activation function. The features generated by convolution layer are transformed into another space by activation function, and the data can be better classified. Note that TanH, Sigmoid, and ReLu are widely applied activation functions. The data is squashed to [-1,1] by TanH function. TanH has a good performance for the features with significant difference, because it can enlarge continuously the effect of features. Besides, TanH is preferred to Sigmoid in practical application due to TanH's mean value is 0. Although the training speed of ReLu function is faster than that of TanH, it is fragile in training and can not use a larger learning rate. Hence, activation layers are equipped with the TanH function in our network.

\subsubsection{Pooling layer}
The pooling layer aims to downsample the feature map. It reduces the number of elements in feature maps, and imports hierarchical structure by increasing the observation window of continuous convolution layers. In our method, max pooling and global average pooling are used. The max pooling which outputs the maximum value in each window of the input feature map is applied to all pooling layers except the fifth one in horizontal/vertical streams. Max pooling layers use kernels of size $3\times3$ and stride of 2, which is the smallest one to capture the notion of left/right, up/down and center. Global average pooling downsamples the feature maps to 1 via the average pooling and it can replace full connection layer to reduce model parameters. Note that global average pooling is only used in the last pooling layer of horizontal and vertical streams.

\section{Experimental Results}
In this section, we first introduce the image datasets and the image processing tool used in our experiments. Then the super parameters of our proposed model are described in detail. Finally, extensive experiments are conducted to assess the effectiveness and efficacy of our proposed resampling detection scheme. Source code of our proposed algorithm is publicly available on the Github \footnote{https://github.com/zhouantao/Resampling/}.
\subsection{Training dataset and image processing}
In following experiments, the image database published by ALASKA steganography challenge is used as testing data \footnote{https://alaska.utt.fr/}. Such a dataset contains 50000 RAW images of size $3154\times5286$ taken by several different cameras. The main image contents include animals, plants, landscapes and buildings, \emph{etc.}. We use the first 5000 images taken by Canon-EOS100D camera. Each image is divided into subimages of $512\times512$ pixels, and seven central subimages are retained. Then JPEG compression is performed using Q=95-97 for simulating the default compression setting in cameras. As such, 35000 subimages are generated as our original image dataset. The datasets used in all sub-experiments would be generated from such an original dataset. All operations are performed on RGB color images for simulating real image processing. The operations are implemented by the built-in image processing toolbox of MATLAB software.

\begin{figure}
\begin{center}
\includegraphics[scale=0.6]{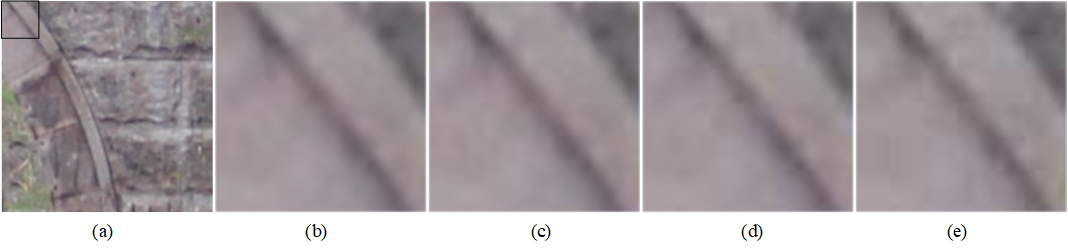}
\caption{Effect of different JPEG quality factors on image. (a) Upsampling 150\% uncompressed images; (b) The black rectangular area in subfigure (a); (c)(d)(e) are JPEG compression versions of (b) with Q=90, 70 and 50, respectively.}
\label{Fig3}
\end{center}
\end{figure}

\subsection{Performance measure and experimental details}
To measure the performance of resampling detection methods, accuracy is employed as evaluation metrics. Our model yields a category label for each image and compares it with the real one. The accuracy is defined as the ratio of correctly classified images' number to the total number of images.

All models in our experiments are implemented using Keras with a single NVIDIA Titan GPU. The network is supervised training using labeled data, and the network weight is initialized by Xavier. The green channel images are fed into the network. We train CNN model using stochastic gradient descent with a batch size of 32 images, momentum \cite{Sutskever2013On} of 0.9, and weight decay of 1e-5. The learning rate is initialized at 0.01. Besides, step decay of learning rate is used in our method. We gradually reduce the learning rate during training. We divide the learning rate by 10 when the validation loss stop improving for three epochs. The number of training epochs is 30 epochs.

\subsection{Resampling detection with recompression}
In this subsection, we set up two different sets of experiments on the detection of resampling in the recompressed image. JPEG is a lossy compression algorithm which can destroy the resampling trace. Figure 3 shows the results when an uncompressed image has undergone JPEG compression with Q=90, 70, 50, respectively. As shown in Figure 3, more image details including the resampling trace are lost and destroyed with the decrease of Q. For Q=50, the blocking artifacts becomes rather obvious.

We first evaluate the performance of our model by specifying the resampling parameter and the Q of JPEG. In the second part, we challenge a difficult task that the resampling parameters and Q are chosen randomly. The performance of the proposed method on different size images is also evaluated.

\subsubsection{On fixed parameters}
In this subsection, we test our proposed method in the recompressed images with different scaling factors and JPEG compression quality factors. We first create a dataset without resampling. We crop the $256\times256$ central area of each image in the original dataset, and then compress it using Q=50-90 with interval of 5. Second, the images in the original dataset are bilinearly scaled by 50\%, 120\% and 150\%, and their central $256\times256$ block image is cut and recompressed with different Qs. As such, 27 sub-databases are generated in terms of the scaling and Qs, and each consists of 35000 unresampled images and corresponding resampled versions. Each database is divided into three parts: 50000 images for training, 10000 images for validation and the rest for testing.

\begin{figure}
\begin{center}
\includegraphics[scale=0.32]{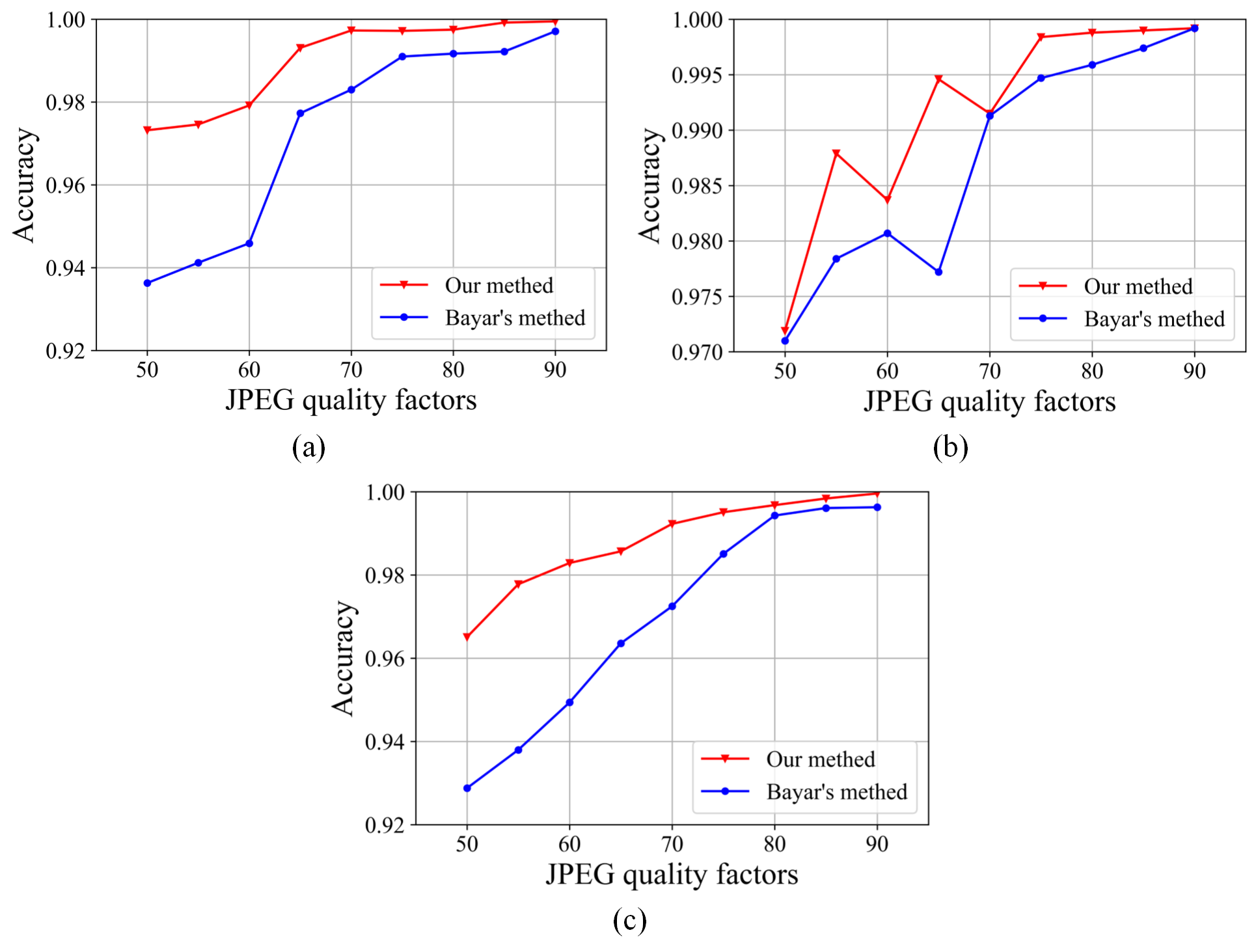}
\caption{Resampling detection accuracy under different post JPEG compression with the scaling factor of (a) 120\%, (b) 150\% and (c) 50\%.}
\label{Fig4}
\end{center}
\end{figure}

In order to estimate the performance of the proposed model, such a model is compared with constrained convolutional neural networks \cite{Bayar2017On,Bayar2018Constrained} using both the same training and testing datasets. The comparison results are illustrated in Figure 4.
The proposed CNN is superior to constrained convolutional neural networks. Our scheme achieves higher detection accuracy in the scenarios of mild upscaling and low quality compression. It can be noted from Figure 4 that our method is similar to Bayar's method \cite{Bayar2017On,Bayar2018Constrained} in performance with high quality post-compression. However, under the post-compression of medium and low quality factors, our method shows a clear advantage. In the case of a QF of 50 and an upsampling of 120, our method improves performance by nearly 4\% compared to Bayar's method; when the downsampling is 50, the performance of our method is also improved by nearly 4\%.

\subsubsection{On random parameters and different image sizes}
In the previous experiments, one certain scaling factor and quality factor are only included in each of databases. Next, we assess the performance of our approach at performing image resampling detection in a more complex scenario where Q and the resampling parameters are chosen arbitrarily. Additionally, the effectiveness of our method is also evaluated under different sizes of images. The original images are first scaled bilinearly with a random scaling factor. The range of upsampling and downsampling factors is respectively from 110\% to 200\% and from 50\% to 90\%, and the interval is 10. Then we retain the central blocks of the resampled images and their original version with different dimensions, i.e., $256\times256$, $128\times128$, $64\times64$. After this, each of the image blocks is compressed using the Q selected arbitrarily in the range of 50-100 with the interval of 10.

\begin{table}
\begin{center}
\small
\tabcolsep3.5pt
\caption{Resampling detection accuracy of different methods in different image sizes.}
\begin{tabular}{lcccccc}
    \specialrule{0.04em}{1pt}{3pt}
    Image size&\multicolumn{2}{c}{$256\times256$} &\multicolumn{2}{c}{$128\times128$} &\multicolumn{2}{c}{$64\times64$}\\
    \cmidrule(r){2-3} \cmidrule(r){4-5} \cmidrule(r){6-7}
    Resampling type&Upsampling &Downsampling &Upsampling &Downsampling &Upsampling &Downsampling \\\specialrule{0.04em}{1pt}{4pt}
    Bayar's method \cite{Bayar2017On,Bayar2018Constrained}	&97.29	&97.24	&95.19	&91.93	&88.19	&80.03\\\specialrule{0em}{1pt}{2pt}
    Proposed method &{\bfseries99.07}	&{\bfseries98.45}	&{\bfseries97.06}	&{\bfseries95.47}	&{\bfseries92.49}	&{\bfseries88.46} \\\specialrule{0.04em}{1pt}{0pt}
\end{tabular}
\end{center}
\end{table}

\begin{table}
\begin{center}
\small
\caption{Parameters of image post-processing operations used in the experiments.}
\begin{tabular}{ll} \hline
        Operation type &Parameters	\\\hline
         Gamma correction(GC)	&$\gamma$: 0.5, 0.6, 0.7, 0.8, 0.9, 1.2, 1.4, 1.6, 1.8, 2.0\\
         Mean filtering(MeanF)	&window size: $3\times3$, $5\times5$, $7\times7$\\
         Gaussian filtering(GF)	&window size: $3\times3$, $5\times5$, $7\times7$, $\sigma$: 0.8-1.6\\
         Median filtering(MedF)	&window size: $3\times3$, $5\times5$, $7\times7$\\
         Wiener filtering(WF)	&window size: $3\times3$, $5\times5$, $7\times7$\\\hline
\end{tabular}
\end{center}
\end{table}

Finally, three databases with different dimensions are created, where each database corresponds to three sub databases, i.e., upsampling-JPEG, downsampling-JPEG and unaltered-JPEG. Each sub database consists of 35,000 images in which 25,000 images are used for training, 5,000 images are used for validation, and the rest part is used for testing. We compare the results of our proposed method and Bayar's method \cite{Bayar2017On,Bayar2018Constrained} under images with different dimensions, as shown in Table 1. The experimental results demonstrate that our model can accurately extract resampling features from images with different resampling factors. Our approach increases the overall classification rate from 88.46\% to 99.07\%. Although the accuracy of our method in $64\times64$ downscaled images is only 88.46\%, it is still far higher than that of Bayar's method. We can notice that the Bayar's method performance degrades severely when reducing the dimension of image.

\subsection{Resampling detection with post-processing}
In order to verify whether our model can maintain excellent performance in more complex scenarios, we used our CNN to detect image resampling when one manipulation is applied for resampled images, followed by JPEG recompression. We design another set of experiments where each image is first resampled, then it is edited by one manipulations, and finally compressed using different Qs. Five image processing methods and parameters are displayed in Table 2, and they are Gamma correction (GC), Mean filtering (MeanF), Gaussian filtering (GF), Median filtering (MedF) and Wiener filtering(WF).

\begin{table}
\begin{center}
\small
\caption{Accuracy of different resampling detection methods against post-processing.}
\tabcolsep5.0pt
\begin{tabular}{lcccccccccc} \specialrule{0.04em}{1pt}{2pt}
        Post-processing type&\multicolumn{2}{c}{GC} &\multicolumn{2}{c}{MeanF}  &\multicolumn{2}{c}{GF} &\multicolumn{2}{c}{MedF} &\multicolumn{2}{c}{WF} \\
        \cmidrule(r){2-3} \cmidrule(r){4-5} \cmidrule(r){6-7} \cmidrule(r){8-9} \cmidrule(r){10-11}
         Resampling type&Up &Down &Up  &Down &Up &Down &Up &Down &Up &Down\\\specialrule{0.04em}{1pt}{4pt}
         Bayar's method \cite{Bayar2017On,Bayar2018Constrained}	&96.50	&95.46	&81.89	&67.03	&85.32	&73.41	&82.39	&67.91	&84.81	&71.14\\
         Proposed method &{\bfseries97.85}	&{\bfseries97.37}	&{\bfseries86.69}	&{\bfseries73.88} &{\bfseries89.14}	&{\bfseries84.26}
          &{\bfseries87.90}	&{\bfseries77.12} &{\bfseries90.45}	&{\bfseries84.46}\\\specialrule{0.04em}{2pt}{0pt}
\end{tabular}
\end{center}
\end{table}

The original images are first resampled with a random scaling factor within 110\%-200\%, 50\%-90\% for upscaling and downscaling, respectively.
The range of upsampling and downsampling factors is respectively from 110\% to 200\% and from 50\% to 90\%, and the interval is 10. Then the resampled image blocks and their corresponding unaltered versions are processed by five operations listed in Table 2, and the operation parameters are selected randomly and uniformly from the set. After this, each manipulated image block is compressed using the selected Q arbitrarily in the range of 50-100 with the interval of 10 and then the central $256\times256$ block is retained. Finally, five databases are built which are GC, MeanF, GF, MedF and WF.

Table 3 depicts the detection accuracy for the proposed method and Bayar's method \cite{Bayar2017On,Bayar2018Constrained} when one manipulation is applied for resampled images. From Table 3, the performances of our and Bayar's methods have different extent of reduction against different post-processing operations. Noticeably, our proposed method is superior to Bayar's method. For upsampled images, median filtering has a greater impact on the detection results of our model with only 87.90\% accuracy, while gamma correction has a smaller impact and with an accuracy rate of 97.85\%. Both values have significantly higher than 81.89\% and 96.50\% of Bayar's method. Although the influence of post-processing operation on downsampling is greater than that on upsampling, our proposed method results are much better than the results of Bayar's method. The detection rates of our approach for each type of manipulation are typically greater than 77\% except for the MeanF operation images which is detected with an accuracy of 73.88\%. Such experimental results show that our scheme can still detect whether the JPEG image with post-processing has been resampled.

\subsection{Estimation of resampling parameters}
In the above experiments, the proposed method can effectively detect the resampling features in the recompressed images with different scaling factors and JPEG quality factors. In this subsection, in order to test the detection capability via resampling parameters, a multiclass classifier is performed for the blind estimation of scaling factors.

Similarly to the previous set of experiments, 35000 image blocks of size $512\times512$ from the original dataset are used to build the database. The images are respectively sampled using bilinear interpolation with fifteen different scaling factors, i.e., 50\%, 60\%, 70\%, 80\%, 110\%, 120\%, 130\%, 140\%, 150\%, 160\%, 170\%, 180\%, 190\% and 200\%. And then the central $256\times256$ blocks of the sampled images are retained to form the final database. Next the generated blocks of size $256\times256$ pixels are compressed using the Q randomly selected in the range of 50-100 at the interval of 10. Eventually, 15 different databases are created. 25000, 5000, 5000 image samples are used for the model training, validation and testing, respectively.

\begin{table}
\begin{center}
\small
\caption{Confusion matrix for identifying resampling parameters using the proposed method. The relative numbers of correct and incorrect classifications are shown in percent. Tru and Pre denote the true and predicted labels, respectively.}
\tabcolsep4.0pt
\begin{tabular}{cccccccccccccccc} \specialrule{0.04em}{1pt}{3pt}
        Tru$\diagdown$Pre&50  &60 &70 &80 &90 &110  &120 &130 &140 &150 &160  &170 &180 &190 &200\\\specialrule{0.04em}{1pt}{4pt}
         50 &{\bfseries91.56}	&4.32	&2.54	&0.76	&0.4	&0.14	&0.02	&0.1	&0.02	&0	&0.04	&0	&0.04	&0.06	&0\\
         60	&4.04	&{\bfseries93.2}	&0.96	&0.9	&0.26	&0.18	&0.08	&0.04	&0.04	&0.12	&0.02	&0.04	&0.04	&0.08	&0\\
         70 &2.3	&1.5	&{\bfseries93.72}	&1.28	&0.7	&0.1	&0.1	&0.06	&0.08	&0.02	&0.02	&0.02	&0.06	&0.02	&0.02\\
        80 &0.62	&0.86	&1.5	&{\bfseries95.46}	&0.98	&0.32	&0.02	&0.02	&0.1	&0.04	&0.02	&0.04	&0	&0.02	&0\\
         90 &0.36	&0.3	&0.84	&0.88	&{\bfseries96.9}	&0.26	&0.12	&0.12	&0.08	&0.02	&0.02	&0.02	&0.02	&0.06	&0\\
         110 &0.08	&0.3	&0.16	&0.46	&0.46	&{\bfseries97.12}	&0.56	&0.26	&0.14	&0.14	&0.02	&0.06	&0.06	&0.14	&0.04\\
         120 &0	&0	&0.1	&0.14	&0.06	&0.58	&{\bfseries98.38}	&0.38	&0.06	&0.1	&0.04	&0.06	&0.04	&0.04	&0.02\\
         130 &0.08	&0.02	&0.02	&0	&0.06	&0.14	&0.24	&{\bfseries98.86}	&0.1	&0.16	&0.08	&0.08	&0.1	&0.04	&0.02\\
         140 &0.06	&0.02	&0.08	&0.18	&0.08	&0.18	&0.12	&0.42	&{\bfseries97.58}	&0.46	&0.24	&0.34	&0.1	&0.06	&0.08\\
         150 &0.04	&0.08	&0.02	&0	&0.06	&0.08	&0.08	&0.1	&0.24	&{\bfseries98.64}	&0.08	&0.32	&0.06	&0.14	&0.06 \\
         160 &0	&0	&0	&0.02	&0	&0.02	&0.06	&0.06	&0.14	&0.14	&{\bfseries98.92}	&0.06	&0.36	&0.12	&0.1\\
         170 &0.04	&0.02	&0.02	&0.02	&0	&0.06	&0.02	&0.02	&0.08	&0.18	&0.12	&{\bfseries98.96}	&0.12	&0.24	&0.1\\
         180 &0.02	&0.02	&0	&0.04	&0.04	&0	&0.06	&0.04	&0.06	&0	&0.34	&0.06	&{\bfseries99}	&0.18	&0.14 \\
         190 &0.02	&0	&0	&0.02	&0.02	&0.02	&0.02	&0.08	&0.1	&0.08	&0.08	&0.12	&0.16	&{\bfseries98.98}	&0.3\\
         200 &0.02	&0	&0	&0	&0.02	&0	&0.02	&0	&0	&0.02	&0.02	&0.08	&0.06	&0.18	&{\bfseries99.58}\\\specialrule{0.04em}{2pt}{0pt}
\end{tabular}
\end{center}
\end{table}

Table 4 shows the confusion matrix achieved by the proposed CNN. Actuals belong on the side of the confusion matrix and predictions are across the top. The accuracies for the proposed CNN with different scaling factors are greater than 91\% and the scaling factor of 200 performs the best results of 99.58\%. For the classifier some images with different scaling factors, such as the scaling factors of 60 and 70, seem to be difficult to distinguish from the scaling factor of 50.

\subsection{Image splicing detection with resampling}
Resampling is a common operation used in image splicing. Here we evaluate the feasibility of our model in image splicing detection. In this experiment, $64\times64$ and $96\times96$ are chosen as a small size to detect resampling. Large size images can be used with $96\times96$ pixels model, otherwise $64\times64$ pixels.

In order to train the variant CNN, we create a dataset using the original dataset. These image blocks are first cut to $64\times64$ and $96\times96$ pixels subblocks, and then 300,000 subblocks are randomly selected, followed by resampling and JPEG recompression, respectively. Scaling factors and quantity factors are randomly chosen in the range of 0.5 to 2 at the interval of 0.01 and the range of 50 to 100 at the interval of 1. Finally, we create two datasets of different image sizes, each consists of 900,000 images (300,000 upsampled images, 300,000 downsampled images and 300,000 unresampled images). The datasets are divided into three parts: 80\% of the images are used for training, 10\% are used for validation, and the

\begin{table}[H]
\begin{center}
\caption{Average detection accuracy for upsampling and downsampling on image blocks with size of $64\times64$ and $96\times96$ pixels.}
\begin{tabular}{ccc} \specialrule{0.04em}{0pt}{1pt}
        Image size     &Upsampling  &Downsampling\\\specialrule{0.04em}{1pt}{3pt}
        $64\times64$ &89\% &77\%\\
        $96\times96$ &92\% &83\%\\\specialrule{0.04em}{1pt}{0pt}
\end{tabular}
\end{center}
\end{table}
\begin{figure}[!htbp]
\begin{center}
\includegraphics[scale=0.44]{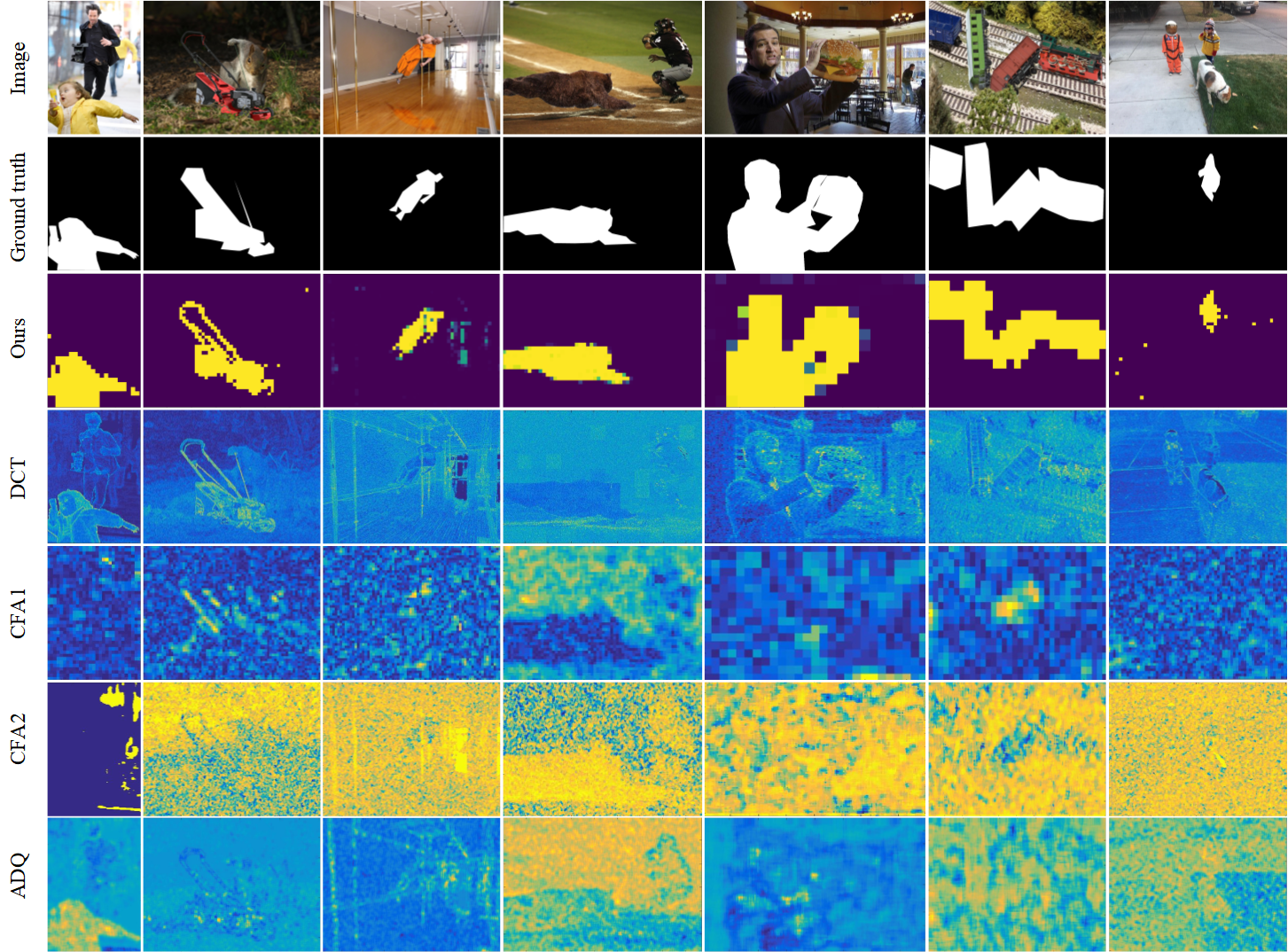}
\caption{Examples from the wild dataset. From top to bottom: forged image, ground truth, results from the five methods: Ours, DCT\cite{Ye2007Detecting}, CFA1\cite{Dirik2009Image}, CFA2\cite{Ferrara2012Image}, ADQ\cite{Amerini2014Splicing}.}
\label{Figure5}
\end{center}
\end{figure}

\noindent rest part is used for testing. Two binary classifier are trained to deal with upsampling and downsampling respectively.

The resampling detection results are shown in Table 5. From Table 5, we can observe when the image size is $96\times96$, the detection accuracy of upsampling is 92\%, and that of downsampling is 83\%. When the image size is reduced to $64\times64$, the accuracy of upsampling classification reaches 89\%, which is already an acceptable result. However, the accuracy of downsampling is only 77\%.

The wild forensics dataset \cite{Huh2018Fighting} is applied to verify the effectiveness of the proposed approach using $64\times64$ pixels in image splicing detection. The tampered areas of the test images are resampled by unknown parameters and compressed by JPEG. Figure 5 shows the results of our and four prior splicing detection methods, which include DCT \cite{Ye2007Detecting}, CFA1 \cite{Dirik2009Image}, CFA2 \cite{Ferrara2012Image}, ADQ \cite{Amerini2014Splicing}. Our method successfully detects and locates the forged region in some images that are not well detected by the other methods. This indicates that the resampling feature is helpful for image splicing detection. However, such a method can only detect the resampled part of the forged image. If the forged part has not been resampled or the image has been processed by a variety of post-processing such as sharpening and filtering, our method will not work.

\begin{table}
\begin{center}
\small
\tabcolsep18pt
\caption{Accuracy of the proposed model on unseen images. The model is trained on ALASKA and tested on both ALASKA and Dersern datasets.}
\begin{tabular}{lcccccc}
    \specialrule{0.04em}{1pt}{3pt}
    &\multicolumn{3}{c}{upscale150} &\multicolumn{2}{c}{upscale110-200} \\
    \cmidrule(r){2-4} \cmidrule(r){5-6}
    &Q=85 &Q=75 &Q=65 &$256\times256$ &$128\times128$ \\\specialrule{0.04em}{1pt}{4pt}
    ALASKA 	&99.94	&99.84	&99.46	&99.07	&98.45	\\\specialrule{0em}{1pt}{2pt}
    Dersern &90.30	&86.03	&80.40	&85.25	&83.41 \\\specialrule{0.04em}{1pt}{0pt}
\end{tabular}
\end{center}
\end{table}

\subsection{Testing on unseen dataset}
Generalization ability, which is the performance of trained models on unknown data, is an important evaluation index for CNN. Here we test the model trained by ALASKA image database on another new dataset. The models trained in Section 3.3.1 (150\% sampling factor, Q=65, 75, 85) and Section 3.3.2 (upsampling operation, image size $256\times256$ and $128\times128$) are tested. The Dersern database \cite{Thomas2010Dresden}, which is a benchmark dataset in image forensics research community, is used as the new test dataset. It contains 16960 JPEG images taken by 27 cameras. We process the Dersern dataset according to the same image processing method as that in the subsection 3.3.1. Finally, 14 datasets are created and each contains 33,920 images, half of which are resampled ones.

As shown in Table 6, the results discover that the detection accuracy on Dersern keeps above 80\% but lower than that on ALASKA, especially for low Qs. For example, the accuracy reduces from 99.94\% to 90.30\% at Q=85, and drops from 99.46\% to 80.46\% at Q=65. This may attribute to the too single source of the training images.

\subsection{Effectiveness validation of model architecture}
The structure of CNN is related to the final detection accuracy. In this subsection, we use a series of experiments to discuss the rationality and effectiveness of our model structure. Three parts of the proposed model have been considered, including the noise extraction layers, horizontal stream or vertical stream and the activation functions. The sub-database (gamma correction, down-sampling) generated in Section 3.4 is utilized to accomplish the experiment here.

\subsubsection{Effectiveness of noise extraction layers}
As described in Section 2.2, two low-order highpass filters are used to extract noise maps in our proposed method. A set of experiments are conducted to prove the effectiveness of noise extraction layers in this section. We consider the three different cases about the noise extraction layer, including

\begin{figure}
\begin{center}
\includegraphics[scale=0.25]{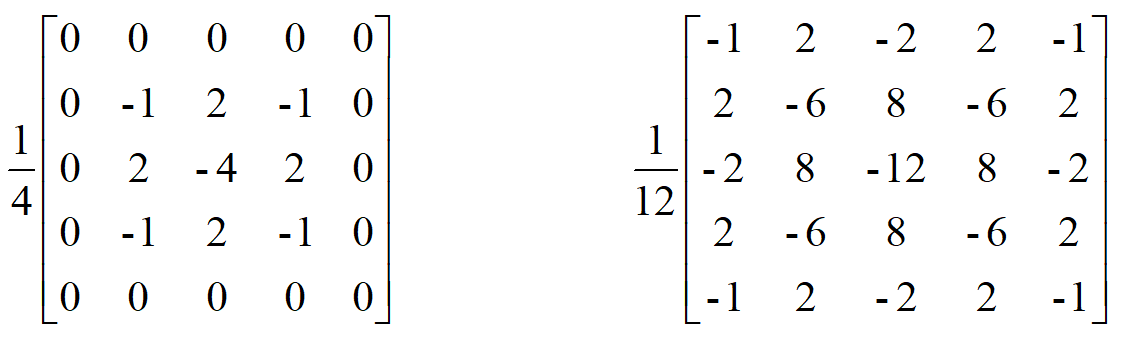}
\caption{Two high-order highpass filters.}
\label{Fig6}
\end{center}
\end{figure}
\begin{figure}[!htbp]
\begin{center}
\includegraphics[scale=0.55]{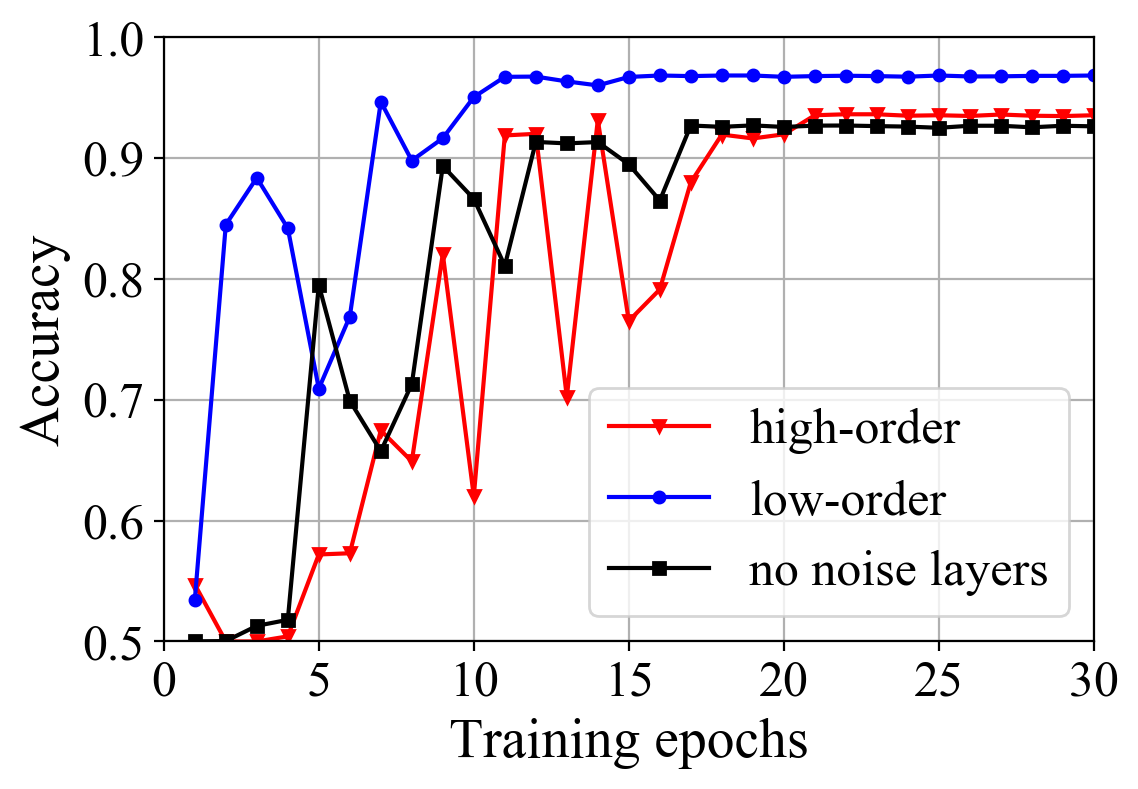}
\caption{Comparison of different settings in the noise extraction layers. Here, high/low-order denote the filters illustrated in Figures 2 and 6, respectively.}
\label{Fig7}
\end{center}
\end{figure}

\noindent using two high-order highpass filters (denoted as ``high-order") from Figure 6, using two low-order highpass filters that used in our method (``low-order") and removing the noise extraction layer (``no noise layer").

The experimental results are shown in Figure 7. We can observe that low-order highpass filter achieves the best performance, and its accuracy is distinctively higher than other methods. The accuracy of high-order highpass filter is slightly higher than removing noise extraction layer.

\subsubsection{Performance of horizontal/vertical and interleaved streams}
In order to assess the performance of horizontal stream and vertical stream, four cases are studied, which are only using horizontal stream, only using vertical stream, using horizontal and vertical streams without interleaved stream, and using horizontal and vertical streams with interleaved stream. Table 7 depicts that the performance of our model is superior to other model, and the accuracy is 97.37\%. The accuracies of horizontal and vertical streams models are 95.52\% and 96.25\%, respectively. Both methods are significantly lower than that in the other two methods. The primary reason is that only one directional feature is considered. The accuracy rate typically greater than 97\% when the horizontal and vertical features are considered simultaneously. Performance of the model with interleaved stream is better than that without interleaved stream, since the directional correlation is further strengthened. Thus, the proposed dual-stream architecture and the interleaved stream strategy could improve the resampling detection performance.

\begin{table}
\begin{center}
\caption{Detection accuracy for upsampling and downsampling of four models.}
\begin{tabular}{ll} \specialrule{0.04em}{0pt}{2pt}

        Different models &Accuracy  \\\specialrule{0.04em}{1pt}{4pt}
        Horizontal stream &95.52\%\\
        Vertical stream &96.25\%\\
        Two streams without interleaved stream &97.04\%\\
        Two streams with interleaved stream &{\bfseries97.37}\%\\\specialrule{0.04em}{2pt}{0pt}
\end{tabular}
\end{center}
\end{table}
\begin{figure}[H]
\begin{center}
\includegraphics[scale=0.55]{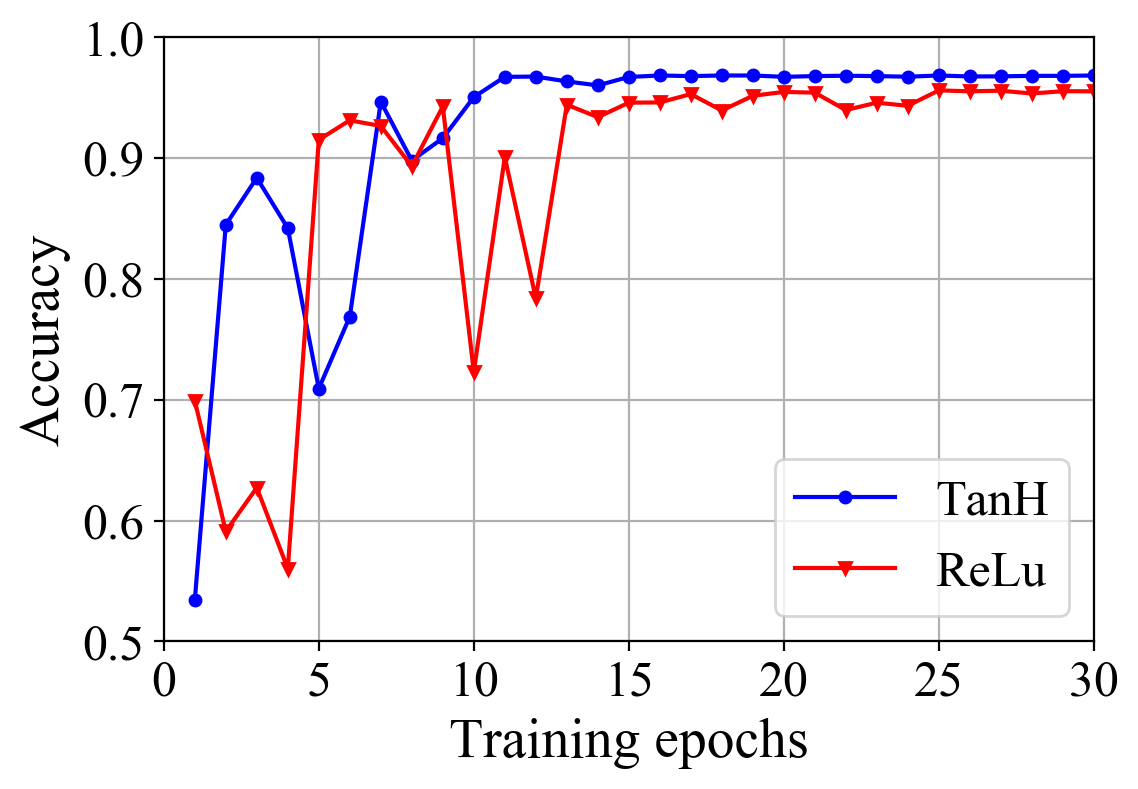}
\caption{Comparison of different activation functions.}
\label{Fig8}
\end{center}
\end{figure}

\subsubsection{Selection of activation function}
Selection of activation function is an important part in the design of convolutional neural networks and different activation functions have great influence on network performance. In order to evaluate the effectiveness of the proposed model with different activation functions, two commonly used activation functions, TanH and ReLu, are chosen. From Figure 8, we can notice that TanH is more suitable for our model. The performance of a network with TanH is better than that of network with ReLu. Furthermore, TanH has better stability and faster convergence than ReLu.

\section{Conclusion}
In this paper, we presented a new deep learning-based method to detect resampling in recompressed images. A noise extraction layer is used to extract noise residual and a dual-stream CNN is proposed to extract resampling features from noise maps in different directions. The experimental results show that the proposed method could not only detect the resampling of recompressed images, but also achieve excellent robustness against some additional post-processing operations. Moreover, we extensively apply the global resampling detection method to resampling parameter estimation and image splicing detection. It should be noted that the proposed method can not detect resampling traces if the image has undergone complex post-processing or anti-forensic manipulations. In the future, we would extend our work to identify more types of resampling in the existence of complex operation chain and anti-forensics attacks.

\section*{Acknowledgments}
This work was supported in part by the National Natural Science Foundation of China (grant nos. 61401408, 61772539, 11571325) and the Fundamental Research Funds for the Central Universities (grant nos. CUC2019B021, 3132017XNG1715).

\section*{Conflict of interest}
All authors declare no conflicts of interest in this paper.

\end{document}